\documentclass[lettersize,journal]{IEEEtran}
\usepackage{amsmath,amsfonts}
\usepackage{algorithmic}
\usepackage{array}
\usepackage[caption=false,font=normalsize,labelfont=sf,textfont=sf]{subfig}
\usepackage{textcomp}
\usepackage{stfloats}
\usepackage{url}
\usepackage{verbatim}
\usepackage{graphicx}
\usepackage{booktabs} 
\usepackage{pifont} 
\usepackage{tabularx} 
\usepackage{float} 
\usepackage[table]{xcolor} 
\usepackage{makecell}
\usepackage{orcidlink}
\usepackage{multirow}

\newcommand{\eg}{\textit{e.g.}}
\newcommand{\ie}{\textit{i.e.}}

\usepackage{color}

\newcommand{\shiqi}[1]{\textcolor{black}{#1}}
\def\BibTeX{{\rm B\kern-.05em{\sc i\kern-.025em b}\kern-.08em
    T\kern-.1667em\lower.7ex\hbox{E}\kern-.125emX}}
\usepackage{balance}
\begin{document}
\title{AvatarBack: Back-Head Generation for Complete 3D Avatars from Front-View Images}
\author{ Shiqi Xin\orcidlink{0009-0000-1481-9090}, ~Xiaolin Zhang\orcidlink{0000-0001-7303-5712}, Yanbin Liu\orcidlink{0000-0003-4724-8065}, Peng Zhang\orcidlink{0000-0001-6794-7352}, Caifeng Shan\orcidlink{0000-0002-2131-1671} 
\thanks{
Corresponding author: Xiaolin Zhang, Caifeng Shan.

Shiqi Xin, Xiaolin Zhang are with the College of Electrical Engineering and Automation, Shandong University of Science and Technology, Qingdao 266590, China. (e-mai1: shiqi.xin@sdust.edu.cn, Xiaolin.Zhang@sdust.edu.cn).

Yanbin Liu is with the Department of Data Science and Artificial Intelligence, Auckland University of Technology, Auckland 1010, New Zealand. (e-mail: yanbin.liu@aut.ac.nz).

Peng Zhang is with the College of Computer Science and Engineering, Shandong University of Science and Technology, Qingdao 266590, China. (e-mail: pengzhang\_skd@sdust.edu.cn).

Caifeng Shan is with the State Key Laboratory for Novel Software Technology and the School of Intelligence Science and Technology, Nanjing University, Nanjing 210023, China. (e-mail: caifeng.shan@gmail.com).
}
}

\maketitle

\begin{abstract}
Recent advances in Gaussian Splatting  have significantly boosted the reconstruction of head avatars, enabling high-quality facial  modeling by representing an 3D avatar as a collection of 3D Gaussians.
However, existing methods predominantly rely on frontal-view images, leaving the back-head poorly constructed. This leads to geometric inconsistencies, structural blurring, and reduced realism in the rear regions, ultimately limiting the fidelity of reconstructed avatars. 
To address this challenge, we propose \textbf{AvatarBack}, a novel \textit{plug-and-play} framework specifically designed to reconstruct complete and consistent 3D Gaussian avatars by explicitly modeling the missing back-head regions. AvatarBack integrates two core technical innovations,~\ie, the \emph{Subject-specific Generator (SSG)} and the \emph{Adaptive Spatial Alignment Strategy (ASA)}.  The former leverages a generative prior to synthesize identity-consistent, plausible back-view pseudo-images from sparse frontal inputs, providing robust multi-view supervision. 
To achieve precise geometric alignment between these synthetic views and the 3D Gaussian representation, the later employs learnable transformation matrices optimized during training, effectively resolving inherent pose and coordinate discrepancies. 
Extensive experiments on NeRSemble and K-hairstyle datasets, evaluated using geometric, photometric, and GPT-4o-based perceptual metrics, demonstrate that AvatarBack significantly enhances back-head reconstruction quality while preserving frontal fidelity. Moreover, the reconstructed avatars maintain consistent visual realism under diverse motions and remain fully animatable. 
\end{abstract}

\begin{IEEEkeywords}
3D Head Reconstruction, Gaussian Splatting, Generative Adversarial Networks. 

\end{IEEEkeywords}
 
\section{Introduction}

\begin{figure}[t]
\centering
\includegraphics[width=1.0\linewidth]{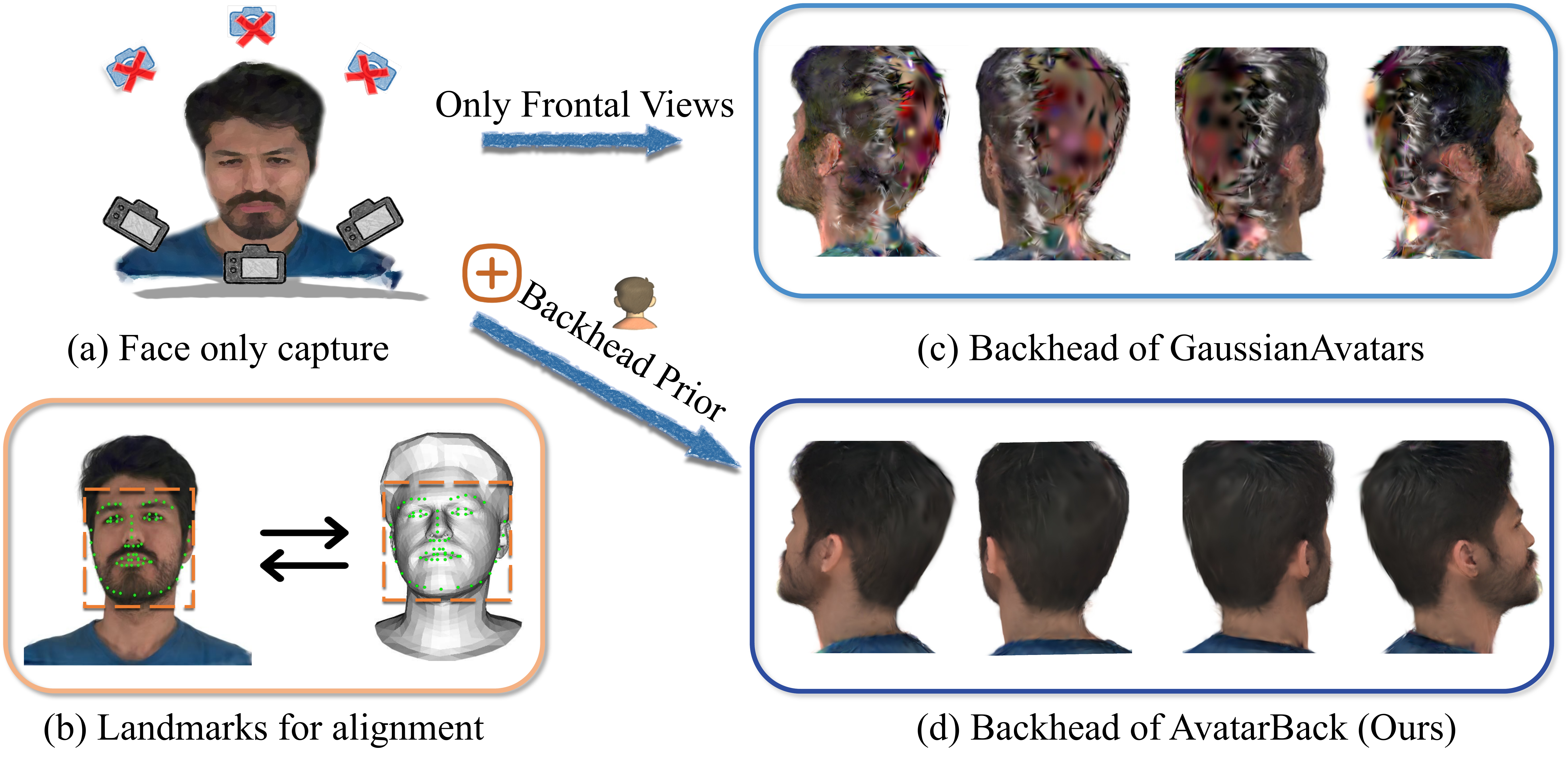}
\caption{\label{fig:back_issue}
Motivation for our AvatarBack. 
(a) Current datasets are limited to frontal views, a result of expensive capture processes.
(b) Existing methods, which rely on facial landmarks for alignment, cannot be applied to the back of the head.
(c) Head reconstruction techniques with  Gaussian splatting~\eg, GaussianAvatars~\cite{Qian2024GaussianAvatars}, fail to reconstruct the backhead regions.
(d) Our proposed method reconstructs the back of the head while maintaining competitive frontal face quality.
}
\vspace{-15pt}
\end{figure}

\IEEEPARstart{3}{D} avatar head reconstruction is an important topic in computer vision and computer graphics, with applications to virtual humans, digital entertainment, virtual reality, and human-computer interaction.
Typical approaches use parametric mesh models to compactly and interpretably represent facial geometry,~\eg, 3DMM~\cite{blanz2023morphable}, FLAME~\cite{Li2017FLAME} and DECA~\cite{Feng2021deca}.
These methods suffer from limitations in depicting fine-grained details of target heads.
Recently, 3D Gaussian Splatting (3DGS)~\cite{kerbl20233d} has emerged as an efficient alternative to achieve both  real-time rendering and photo-realistic reconstructions by modeling 3D points as differentiable Gaussian distributions. 
Building on 3DGS, methods,~\eg, GaussianAvatars~\cite{Qian2024GaussianAvatars}, SplattingAvatar~\cite{Shao2024SplattingAvatar}, FlashAvatar~\cite{Xiang2024FlashAvatar}, SurfHead~\cite{lee2025surfhead}, and PSAvatar~\cite{zhao2024psavatar}, combine mesh priors, rigging, and Gaussian primitives to generate photorealistic and animatable head avatars.
These avatars capture dynamic expressions and detailed appearance, leveraging the flexibility and efficiency of Gaussian-based representations. 
However, despite these impressive advances, a persistent challenge in 3D head reconstruction, particularly for 3DGS-based methods~\cite{Qian2024GaussianAvatars,Shao2024SplattingAvatar,Xiang2024FlashAvatar,lee2025surfhead,zhao2024psavatar,zhong2025avatarmakeup}, arises from their reliance on frontal or limited-view datasets. This limitation stems mainly from two factors: 1) acquiring full 360$^\circ$ multi-view data is expensive and time-consuming (Fig.~\ref{fig:back_issue}(a)), and 2) most frontal reconstruction models are designed around facial detection and landmark alignment, focusing almost exclusively on the visible front (Fig.~\ref{fig:back_issue}(b)). 
As a result, these methods struggle to reconstruct the crucial but under-constrained back-head, often producing incomplete 3D portraits where rear regions are poorly represented or entirely missing. As shown in  Fig.~\ref{fig:back_issue}(c), this leads to unrealistic and visually jarring back-head geometry, which significantly reduces the realism and usability of reconstructed avatars.

Although some recent methods attempt to address missing views, they remain fundamentally ineffective in practice. For example, PanoHead~\cite{Sizhe2023PanoHead} generates full-head appearances from a single frontal image using latent feature synthesis. However, it relies on implicit geometry representations and does not produce explicit representations, making it incompatible with pipelines that require controllable and animatable 3D geometry. Other 3DGS-based methods focus on enhancing facial detail and expression control but leave the back-head under-constrained due to the absence of direct supervision~\cite{Shao2024SplattingAvatar,Xiang2024FlashAvatar,zhao2024psavatar}. As a result, no existing approach effectively combines plausible back-head completion with explicit, high-fidelity, and fully animatable 3D reconstruction.

Motivated by this gap, we tackle the challenging task of reconstructing complete 3D Gaussian head avatars when back-view regions are missing from the input data. Our key idea is to leverage generative models to synthesize the unobserved back-head regions in a way that remains consistent with the subject’s frontal appearance and hairstyle (Fig.~\ref{fig:back_issue}(d)). While such generative supervision is plausible in image space, integrating it into a point-based 3DGS framework introduces two main challenges: 1) maintaining visual coherence with the frontal face to avoid identity or hairstyle artifacts, and 2) accurately aligning the pseudo-image content with the 3DGS spatial domain to ensure effective and stable reconstruction. Addressing these challenges requires a novel solution that bridges generative supervision and explicit, animatable 3D Gaussian modeling.

To this end, we propose \textbf{AvatarBack}, a novel framework designed to complete the missing back-head regions in 3D Gaussian head avatars. 
AvatarBack integrates two core modules: the \emph{Subject-specific Generator (SSG)} and the \emph{Adaptive Spatial Alignment Strategy (ASA)}. 
SSG leverages a generative prior to synthesize plausible back-view pseudo-images that are consistent with the subject’s frontal appearance and hairstyle, providing crucial supervision signals for the 3DGS model. 
To ensure these synthetic views are effectively utilized, an ASA employs learnable transformation matrices, optimized during training, to align the pseudo-image content accurately with the 3DGS model space. 
Together, these modules form a unified reconstruction framework. AvatarBack achieves significantly more complete and geometrically consistent 3D head avatars, with notable improvements in the rear regions. Meanwhile, it preserves the explicit, animatable, and high-fidelity properties of Gaussian-based methods.

To the best of our knowledge, this work presents the first approach specifically designed to complete the back-head regions in Gaussian-based 3D head avatars. The proposed \textbf{AvatarBack} framework is a \textit{\textbf{plug-and-play}} solution that integrates flexibly into state-of-the-art systems such as GaussianAvatars~\cite{Qian2024GaussianAvatars} and SurfHead~\cite{lee2025surfhead}. We establish a comprehensive evaluation protocol combining geometric, photometric, and perceptual metrics, including a GPT-4o-based scoring method for assessing fine-grained, human-aligned visual quality of the reconstructed back-view. Experiments on the NeRSemble~\cite{Kirschstein2023NeRSemble} and K-hairstyle~\cite{Kim2021KHairstyle} datasets show that AvatarBack significantly improves back-head reconstruction while preserving frontal fidelity. Moreover, the enhanced avatars remain fully animatable and reliably respond to diverse driving signals, consistently delivering high visual realism across a wide range of motions and expressions.

In summary, our main contributions are as follows:
\begin{itemize}
\item We propose AvatarBack, the first framework to complete the back-head regions in Gaussian-based 3D head avatars, combining generative supervision with explicit 3D Gaussian modeling.
\item We design a plug-and-play solution with two novel components: SSG for synthesizing consistent back-view, and ASA for accurate integration into the 3DGS space.
\item We establish a comprehensive evaluation protocol, including geometric, photometric, and GPT-4o-based perceptual metrics, demonstrating superior reconstruction quality and full animatability on challenging benchmarks.
\end{itemize}

\section{Related Work}
\subsection{3D Gaussian Splatting for Head Avatars}
The recent 3D Gaussian Splatting (3DGS) \cite{kerbl20233d} offers a new paradigm with state-of-the-art, real-time rendering. However, it represents geometry as an unstructured Gaussian cloud, which lacks the explicit topology for animation and editing. A dominant strategy is to impose structure by anchoring the Gaussians to an explicit mesh, enabling coherent control.
For example, SplattingAvatar \cite{Shao2024SplattingAvatar} embeds Gaussians within a mesh using barycentric coordinates, enabling mesh-driven animation. Similarly, GaussianAvatars \cite{Qian2024GaussianAvatars}, HERA \cite{Cai2025HERA}, SVG-Head \cite{sun2025svghead}, and MeGA \cite{Wang2025MeGA} all leverage an underlying surface (explicit or implicit) to provide topological consistency and control, allowing for high-fidelity rendering and coherent deformation. SurfHead \cite{lee2025surfhead} further refines this by using 2D Gaussian surfels constrained on a deforming mesh, adeptly handling extreme poses.
Another significant research thrust focuses on creating compact and parametrically controllable models. This is often achieved by integrating Gaussians with established parametric head models. For example, GPHM \cite{xu2024gphm} and NPGA \cite{Giebenhain2024NPGA} leverage expression and pose parameters to directly drive the attributes of the Gaussians, enabling semantic control. Others achieve compactness through learned representations, such as reduced Gaussian blendshapes \cite{Li2025RGBAvatar}, graph neural networks \cite{Wei2025GraphAvatar}, or efficient latent embeddings in texture space \cite{Li2025TeGA} or high-dimensional spaces \cite{serifi2025hypergaussians}, all aiming for efficient animation and reduced model size.
Furthermore, recent efforts have pushed the boundaries of realism and interactivity. A key focus is relightability, where methods such as BecomingLit \cite{schmidt2025becominglit}, HRAvatar \cite{Zhang2025HRAvatar}, and LightHeadEd \cite{manu2025lightheaded} disentangle intrinsic surface properties from illumination, allowing avatars to be realistically rendered under novel lighting conditions. Specialized components, such as dynamic and plausible hair, have also been addressed by dedicated hybrid models \cite{liao2024hhavatar, Zheng2025GroomLight}. Moreover, the field is moving towards greater user control, enabling direct textural editing \cite{Zhang2025FATE} or creating generative, interactive avatars \cite{Yu2025GAIA}.
While these methods perform well in supervised areas, they often lack sufficient guidance in sparse regions, such as the back of the 3D avatar head. As a result, recovering complete head geometry remains a challenging task.

\begin{figure*}[t]
\centering
\includegraphics[width=7 in]{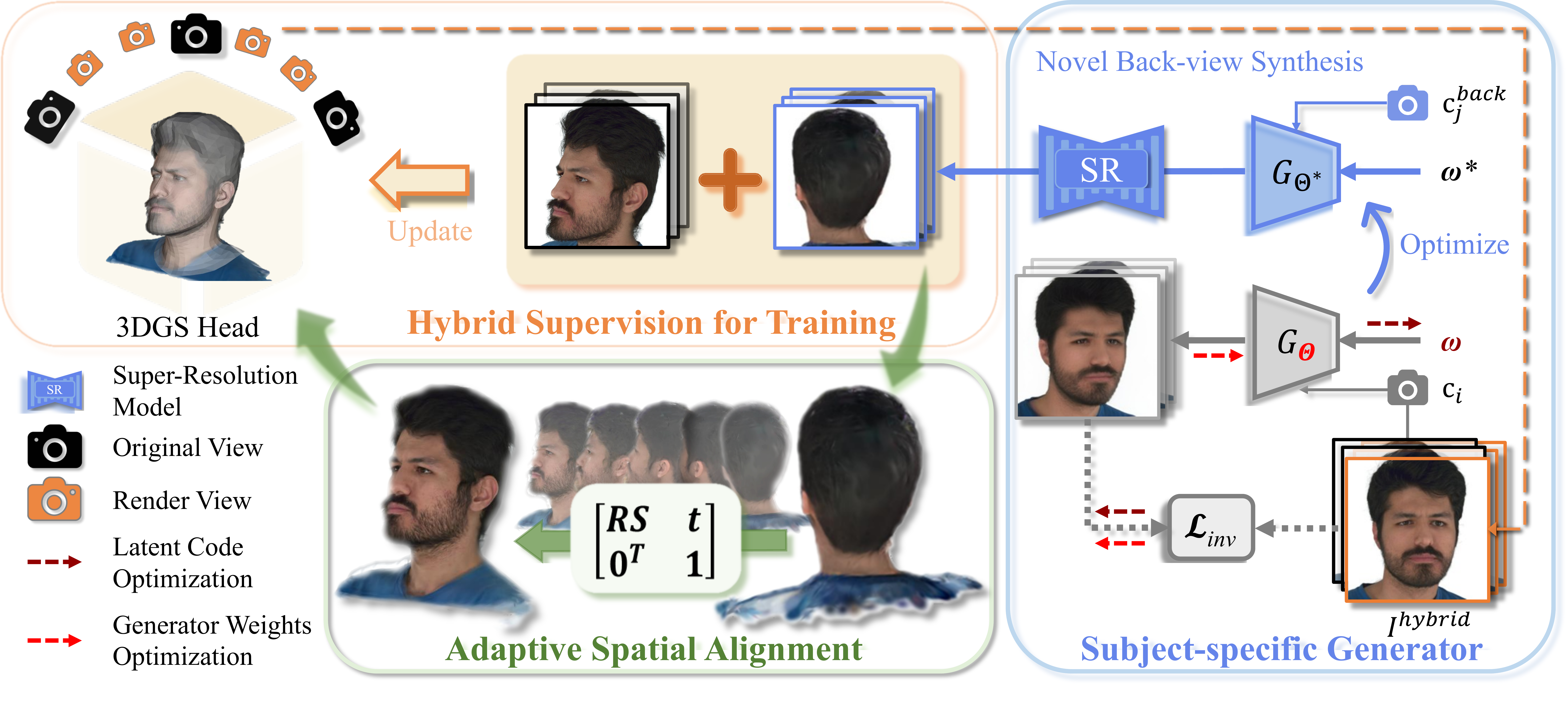}
\vspace{-5pt}
\caption{
Overview of the AvatarBack framework. The framework leverages a \textit{Subject-specific Generator} to produce pseudo back-head views via generator optimization $(\mathbf{w}, \Theta)$. These views are adaptively aligned during reconstruction through an \textit{Adaptive Spatial Alignment Strategy}  module and jointly supervised with real frontal images. This hybrid training strategy enables identity-consistent and complete \textit{full-head }reconstructions, effectively compensating for missing supervision in invisible regions. 
}
\label{fig:method}
\vspace{-15pt}
\end{figure*}

\subsection{3D-aware GANs}
Generative Adversarial Networks (GANs) \cite{goodfellow2014generative} have achieved major advances in image generation tasks since their introduction. 
The style-based GAN architecture (StyleGAN)~\cite{Karras2019styleGAN}, in particular, generated high-quality images by disentangling style and content. 
This powerful disentanglement paved the way for advanced latent space editing techniques for various applications \cite{ patsouras2021styleclip, khan2025instaface,zhong2024dreamlcm}.
Recently, GANs have also been applied to the field of 3D modeling. 
3D-GAN \cite{Wu20163DGN} pioneered the application of adversarial training to voxel grids. It synthesized 3D shapes from random noise using convolutional voxel generators and discriminators. However, limitations in voxel resolution hindered the representation of fine details. 
Subsequently, researchers developed GAN frameworks for various 3D representations, such as point clouds \cite{Li2021SPGAN, Yang2022CGAN}, explicit meshes \cite{Pemasiri2022Im2MeshGAN, gao2022get3d}, and implicit fields \cite{Chan2022EG3D,Sun2024NeRFEditor}. 
These advancements significantly enhanced generation quality and diversity.

GAN-based methods have also started showing promise in 3D human head reconstruction. EG3D \cite{Chan2022EG3D} introduced a hybrid architecture combining tri-planes and neural rendering. This approach decouples feature generation from the rendering process. It enables the real-time generation of multi-view consistent, high-resolution images and high-quality geometry. Building on this, PanoHead \cite{Sizhe2023PanoHead} incorporated facial priors and geometric regularization. This allows the generation of complete 360° head models from a single frontal image. It effectively completes the structure and texture of unseen areas, such as the back regions of the 3D avatar head. Furthermore, Pivotal Tuning Inversion (PTI) \cite{Roich2022PTI} introduced an innovative GAN inversion technique. It fine-tunes a pre-trained StyleGAN generator to better match an input image. This enables high-quality reconstruction and editing. Applying PTI to PanoHead further enhances its reconstruction accuracy. 

However, these 3D GAN methods can still produce artifacts, such as view inconsistencies or structural misalignments. This often occurs when handling extreme viewpoints or sparse real image inputs. Moreover, these models typically store 3D information solely as latent variables. This makes it difficult to directly generate drivable 3D human head structures with rich details.

\section{Method}
\label{sec:method}

\subsection{Overall Framework}

Given a training set of multi-view frontal images $\{I_i\}_{i=1}^{N_{ori}}$ for a specific target subject, we follow state-of-the-art methods, namely GaussianAvatars~\cite{Qian2024GaussianAvatars} and SurfHead~\cite{lee2025surfhead}, for 3D head reconstruction. 
The reconstructed 3D head avatar integrates 3D Gaussian Splatting (3DGS) with a parametric FLAME mesh representation. 
The FLAME model provides the underlying geometry and animatable structure, while the 3DGS captures detailed textures to render realistic appearance. 
These two representations are coupled by binding the relative positions of Gaussian kernels near the mesh surface.

Formally, we denote a parametric FLAME mesh as $\mathcal{M}(\boldsymbol{\phi})$, where $\boldsymbol{\phi}$ encodes identity, expression, and pose parameters. 
For each triangle center $p_k$ on the mesh, a Gaussian kernel is attached and parameterized as  
\begin{equation}
\mathcal{G}_k = \{ \boldsymbol{\mu}_k, \mathbf{\Sigma}_k, \mathbf{R}_k, \boldsymbol{\alpha}_k, \mathbf{c}_k \},
\end{equation}
where $\boldsymbol{\mu}_k \in \mathbb{R}^3$ denotes the mean position, $\mathbf{\Sigma}_k \in \mathbb{R}^{3\times3}$ represents the anisotropic scale, $\mathbf{R}_k \in SO(3)$ specifies the rotation, $\boldsymbol{\alpha}_k$ is the opacity, and $\mathbf{c}_k$ is the view-dependent color.  
During animation, the deformation of triangles driven by pose and expression changes is propagated to each Gaussian kernel $\mathcal{G}_k$, ensuring consistent motion and appearance across the 3D avatar.

Following such representation and captured frontal face images, it can produce a photorealistic and temporally stable reconstruction of the visible facial regions in the training views. 
However, the back-head remains unobserved and unsupervised due to view sparsity, resulting in incomplete geometry and texture. 
As shown in Fig.~\ref{fig:method}, we propose \textbf{AvatarBack}, a framework that incorporates synthesized images as pseudo-supervisions to complement the reconstruction of unobserved regions,~\eg, back-head areas. 
The framework introduces a \emph{3DGS-aware closed feedback loop} between reconstruction and generative synthesis. 
Instead of treating 3DGS reconstruction and image synthesis as independent stages, AvatarBack couples them so that: 
1) an initial 3DGS head model provides geometry- and pose-consistent cues to guide plausible back-view synthesis, and  
2) the synthesized views are fed back into the 3DGS pipeline as pseudo-supervision to progressively refine the missing regions. 

To fulfill both purposes, we propose a \textbf{S}ubject-\textbf{s}pecific \textbf{G}enerator (\textbf{SSG}) module for providing pseudo-supervision of the unobserved back-head regions. 
Additionally, an \textbf{A}daptive \textbf{S}patial \textbf{A}lignment (\textbf{ASA}) module aligns the synthesized and captured supervisions with the underlying 3D coordinates.
Specifically, \textbf{SSG} (Sec.~\ref{sec:multi-view-inversion}) uses a hybrid supervision set that combines captured real views and 3DGS-rendered novel views to drive a geometry-conditioned GAN inversion process, producing identity-consistent back-head views even under extreme view sparsity. 
\textbf{ASA} (Sec.~\ref{sec:method-Alignment}) ensures pixel-accurate integration of the generated images into the 3DGS coordinate space via a learnable geometric transformation. 

By enabling feedback between generated pseudo images of unobserved regions and rendered realistic images of the 3DGS head avatars, both components can be mutually reinforced, yielding complete, animatable, and photorealistic 3D head avatars. 
Particularly, the 3DGS model gains dense supervision from realistic, identity-preserving novel views. 
The generated pseudo-images can be further improved by integrating explicit 3DGS geometry for appearance-consistent synthesis.

\subsection{Subject-specific Generator for Unseen Back-head Regions}
\label{sec:multi-view-inversion}

Given a 3DGS reconstruction of a head avatar from a state-of-the-art approach,~\eg, GaussianAvatars and SurfHead, 
the 3DGS head accurately reconstructs the frontal region but leaves the back-head largely unconstrained due to the absence of such views in the training set.  
We utilize a pretrained 3DGAN model,~\ie, PanoHead~\cite{Sizhe2023PanoHead}, to synthesize the unseen structure of the missing regions, and the synthetic images of the unobserved regions are employed as pseudo-supervision for the optimization of back-head regions. 
To meet the fundamental consistency requirements of the synthetic images with the target subject, an inversion of the 3DGAN model is a natural choice. 
However, directly performing 3DGAN inversion on PanoHead produces artifacts in the back-head regions and does not provide effective supervision for the back-head completion task. 

Therefore, instead of directly applying a generic inversion pipeline, which is prone to identity drift under sparse and biased viewpoints, we adopt a feedback loop and introduce a subject-specific generator strategy. 
This strategy explicitly exploits the initialized 3DGS model to strengthen inversion and, in turn, uses the inversion outputs to enhance the 3DGS reconstruction.

Concretely, the Subject-specific Generator consists of two phases: the Hybrid Subject-specific Generator and Novel Back-view Synthesis. 
The Hybrid Subject-specific Generator aims to adapt the hidden weights of PanoHead to a specific subject. 
Instead of using only limited frontal-view images, we adopt a hybrid multi-view inversion strategy that incorporates both the captured real images and the rendered images from the 3DGS avatars. 
As the reconstruction process relies on the supervision of both types of images, the inversion process can therefore benefit from the updated reconstructions.

Specifically, the inversion process can be formulated as an optimization task as in Eq.~\eqref{inversion}:
\begin{equation}\label{inversion}
\begin{split} 
&(\mathbf{\Theta^*}, \mathbf{w^*})=\operatorname*{arg\,min}_{\mathbf{\Theta}, \mathbf{w}} \frac{1}{N}\sum_{i=1}^N \mathcal{L}_{\mathrm{inv}}\!\left(G_{\mathbf{\Theta}}(\mathbf{w}, c_i), I_i\right), \\ &I_i \in \mathcal{I}^{hybrid} = \{ I_i^{\mathrm{ori}} \}_{i=1}^{N_{\mathrm{ori}}} 
\;\cup\; 
\{ I_i^{\mathrm{render}} \}_{i=1}^{N_{\mathrm{render}}},
\end{split}
\end{equation}
where $N = N_{\mathrm{ori}} + N_{\mathrm{render}}$, and $\mathbf{w}$ and $\Theta$ are the latent code and the weights adapted to the target subject, respectively. 
$\{c_i\}$ denotes the calibrated camera pose with respect to the given image $I_i$. 
$\mathcal{L}_{\mathrm{inv}}$ combines pixel-wise, perceptual, and latent regularization losses, enforcing appearance coherence across both captured and 3DGS-rendered views. 
The hybrid dataset $\mathcal{I}^{hybrid}$ mitigates the underconstrained nature of sparse frontal captures, providing diverse geometry-aware cues to the inversion stage. 

Different from the vanilla inversion method, which uses a single frontal-view image to obtain the latent code of a generator, we utilize a hybrid multi-view inversion strategy to obtain an appearance-consistent model across different views. 
For a specific subject, the original captured images $\{ I_i^{\mathrm{ori}} \}_{i=1}^{N_{\mathrm{ori}}}$ normally include only sparse views of the frontal face region. 
By incorporating the rendered images from the updated 3DGS head, the inversion process is constrained to maintain identity and multi-view consistency, improving the details of the generated pseudo-images for back-head regions.

After adapting the optimal latent code $\mathbf{w}^*$ and generator weights $\mathbf{\Theta}^*$ to the target head, we utilize them in the synthesis process of the novel back-head views.  
Given a specific novel view $c_j^{\text{back}}$ of the back-head, we render the pseudo-image $I^{\text{back}}$ by forwarding the adapted generator following Eq.~\eqref{sampling}:  
\begin{equation}\label{sampling}
I^{\text{back}} = G_{\mathbf{\Theta^*}}\!\left(\mathbf{w}^*, c_j^{\text{back}}\right).
\end{equation}
To obtain images of the unobserved back-head regions, $\{c_j^{\text{back}}\}$ are sampled from azimuths $[90^\circ, 270^\circ]$.  
This novel back-view synthesis process can produce a set of synthetic back-head images, and these images maintain strong identity coherence thanks to the participation of the rendered 3DGS images.

Furthermore, the synthesized back-view are refined using the face-oriented super-resolution network~\cite{He2022GCFSR}, which enhances high-frequency details such as hair strands, edges, and contour sharpness.  
These enhanced images $\{\tilde{I}_j^{\text{back}}\}$ form the high-quality pseudo-supervision set used in subsequent 3DGS-based reconstruction, ensuring that fine details are preserved in the final completed 3D head model.

Through this closed-loop process, the inversion stage is no longer an isolated pre-processing step. Instead, it is explicitly \emph{coupled with the 3DGS model}, both benefiting from and contributing to it.  
This design enables reliable, identity-consistent back-view generation even under severe viewpoint sparsity.

\subsection{Adaptive Spatial Alignment}
\label{sec:method-Alignment}
Precise pixel-level alignment between rendered 3DGS head views and generated pseudo-images is essential for supervising the back-head regions.
For instance, if a generated view depicts a back-head hair part at a specific image location, the corresponding rendering from the 3DGS head model must reproduce that detail at the same position and scale. Otherwise, the supervision may introduce inconsistencies or errors. However, due to differences in the coordinate systems of the Gaussian avatar and the generative model, their rendered images are misaligned. 

To resolve this, we introduce an \textbf{A}daptive \textbf{S}patial \textbf{A}lignment (\emph{ASA}) module, which learns a transformation matrix $\mathbf{T} \in \mathbb{R}^{4 \times 4}$ that aligns the coordinate systems of the two models. The transformation is applied to the FLAME mesh vertices $\mathcal{V}(\phi)$, thereby indirectly guiding the 3DGS head to align with the generative model coordinate. This matrix is optimized jointly during training via the following objective:
\begin{equation}
\begin{aligned}
\label{eq:multi_view_transform}
\mathbf{T^*} = &\arg\min_{\mathbf{T}} \sum_{j=1}^{M} 
\mathcal{L}\left(\mathcal{R}(\mathcal{G}(\mathbf{T} \cdot \mathcal{V}(\phi),\; c_j^{\text{back}}),\; I_j^{\text{back}} \right) \\
&+ \mathcal{L}_{\mathrm{FLAME}}(\phi),
\end{aligned}
\end{equation}
where $\mathcal{G}$ is the set of 3D Gaussians, $c_j^{\text{back}}$ denotes a back-view camera pose, and $I_j^{\text{back}}$ is the pseudo back-head image. The function $\mathcal{R}$ denotes the differentiable Gaussian rendering. 

The loss $\mathcal{L}$ enforces photometric consistency between rendered and pseudo images, while $\mathcal{L}_{\mathrm{FLAME}}$ regularizes the alignment to remain faithful to the original FLAME geometry:
\begin{equation}
\mathcal{L}_{\mathrm{FLAME}}(\phi) = \lambda_{\text{flame}} \cdot \left\| \boldsymbol{\phi} - \boldsymbol{\phi}_{\text{orig}} \right\|_2^2,
\label{eq:flame_reg}
\end{equation} 
where $\boldsymbol{\phi}$ and $\boldsymbol{\phi}_{\text{orig}}$ denote the current and initial FLAME parameters, and $\lambda_{\text{flame}}$ controls the regularization strength.

Directly optimizing the entire transformation matrix \( \mathbf{T} \) can easily introduce undesired effects beyond intended scaling, rotation, and translation. It may also lead to coupling between different geometric components, making independent control difficult. Therefore, we decompose \( \mathbf{T} \) into a scale matrix \( \mathbf{S} \), a rotation matrix \( \mathbf{R} \), and a translation vector \( \mathbf{t} \), which are optimized separately. Formally, the transformation matrix $\mathbf{T}$ can be expressed as:
\begin{equation}
\mathbf{T} = \begin{bmatrix}
\mathbf{R}\mathbf{S} & \mathbf{t} \\
\mathbf{0}^\top & 1
\end{bmatrix},
\end{equation}
where $\mathbf{R} \in \mathbb{R}^{3 \times 3}$, $\mathbf{S}\in\mathbb{R}^{3\times 3}$, 
$\mathbf{t}\in\mathbb{R}^{3}$, 
and $\mathbf{0}\in\mathbb{R}^{3}$.

To facilitate optimization, we parameterize the spatial transformation using low-dimensional vectors. Specifically, the transformation consists of a learnable scale vector \( \mathbf{s} \in \mathbb{R}^{3} \), a rotation vector \( \mathbf{r} \in \mathbb{R}^{3} \), and a translation vector \( \mathbf{t} \in \mathbb{R}^3 \). The scale and rotation vectors are converted into matrix form during the forward pass, while the translation vector is directly applied without further transformation.

The scale matrix \(\mathbf{S}\) is defined as:
\begin{equation}
\mathbf{S} = \mathrm{diag}(\mathbf{s}) = 
\begin{bmatrix}
s_1 & 0 & 0 \\
0 & s_2 & 0 \\
0 & 0 & s_3
\end{bmatrix},
\end{equation}
and the rotation matrix \( \mathbf{R} \) is computed using Rodrigues' rotation formula:
\begin{equation}
\mathbf{R} = \mathbf{I} + \frac{\sin \theta}{\theta} [\mathbf{r}]_{\times} + \frac{1 - \cos \theta}{\theta^2} [\mathbf{r}]_{\times}^2,
\end{equation}
where \( \theta = \|\mathbf{r}\| \) is the magnitude of the rotation vector, and \( [\mathbf{r}]_{\times} \in \mathbb{R}^{3 \times 3} \) is the corresponding skew-symmetric matrix:
\begin{equation}
[\mathbf{r}]_{\times} = 
\begin{bmatrix}
0 & -r_3 & r_2 \\
r_3 & 0 & -r_1 \\
-r_2 & r_1 & 0
\end{bmatrix}.
\end{equation}
Here, \( \mathbf{I} \) denotes the \( 3 \times 3 \) identity matrix. The rotation matrix \( \mathbf{R} \) is thus obtained by exponentiating \( [\mathbf{r}]_{\times} \), ensuring a smooth and stable representation during optimization.

\section{Experiments}
\subsection{Implementation Details}

As a plug-and-play algorithm, we verify the proposed method by integrating it with state-of-the-art approaches,~\eg, GaussianAvatars~\cite{Qian2024GaussianAvatars} and SurfHead~\cite{lee2025surfhead}.
We augment the training process with supervision from pseudo-images of the back-head. These pseudo-images are integrated into the pipeline alongside an ASA module, which is co-optimized with the 3DGS head avatar. This unified framework allows for joint optimization using a combined objective over real and spatially-aligned pseudo-images.
For training, real multi-view images of the frontal face are utilized at a resolution of 802$\times$550 pixels, while pseudo-images generated by our proxy module are 512$\times$512 pixels. The entire training process employs the Adam optimizer, with the loss weight for pseudo-real images set to $\lambda=0.01$.
The weight of the FLAME parameters constraint is $\lambda_{\text{flame}} = 0.5$ to enforce geometric consistency. The initial learning rate for scale factor $\textbf{s}$, rotation matrix $\textbf{r}$, translation vector $\textbf{t}$ is 0.005, following a cosine annealing schedule.
All experiments are conducted on a workstation equipped with four NVIDIA RTX 3090 Ti GPUs, each with 24 GB of memory.

\subsection{Experimental Setup}

\begin{figure*}[!t]
    \centering
    \includegraphics[width=\textwidth]{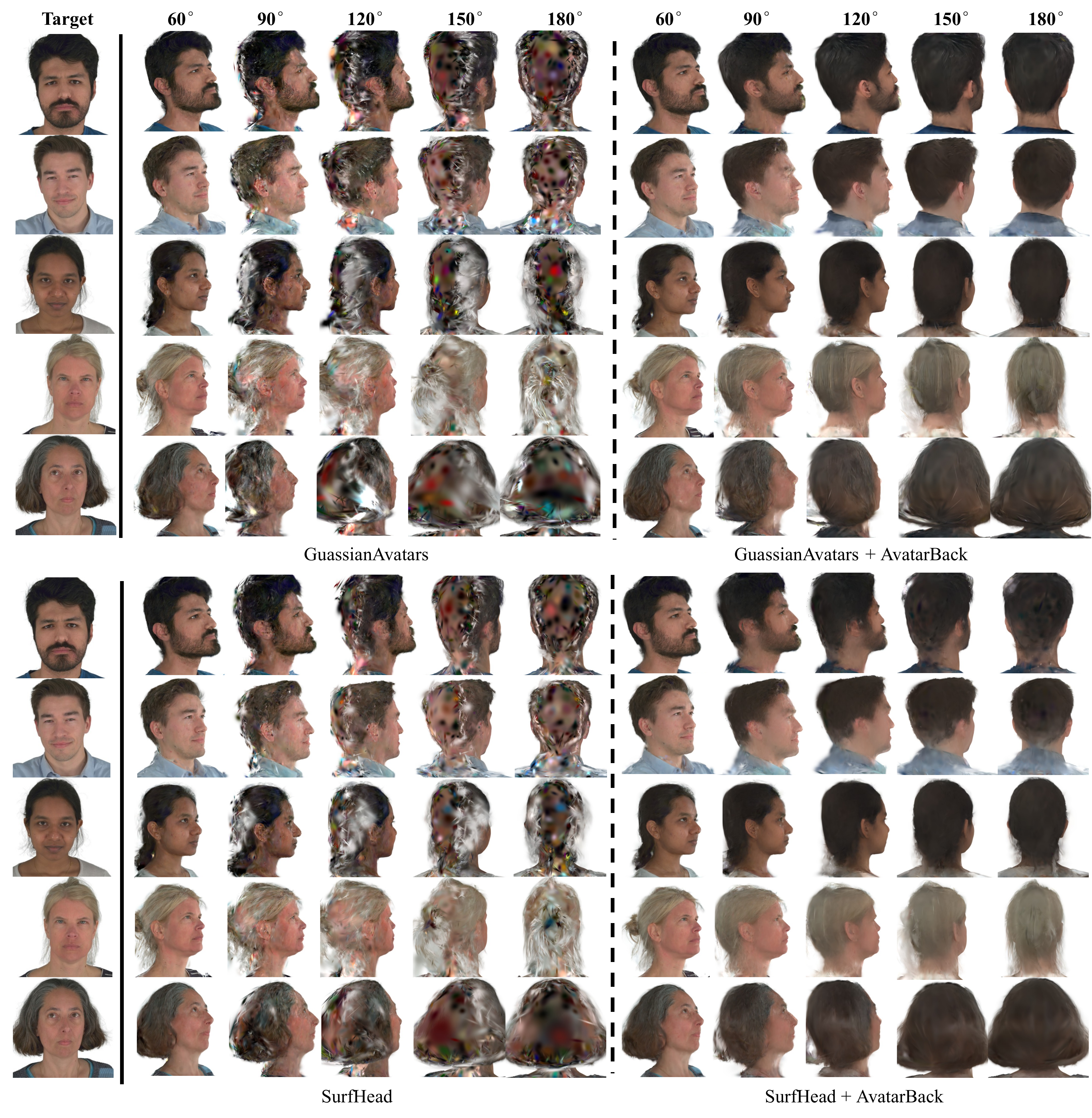}
    \vspace{-20pt}
    \caption{
\shiqi{Qualitative comparison of back-head reconstructions. The first column displays ground-truth images, followed by five rendered views under azimuth angles from $60^\circ$ to $180^\circ$.} 
The proposed AvatarBack method can be integrated into existing models,~\eg, GaussianAvatars and SurfHead, to accurately reconstruct the unseen back and side head regions.}
    \label{fig:backview_comparison_gs}
    \vspace{-15pt}
\end{figure*}

\textbf{Datasets.}
We employ two publicly available datasets, \textbf{NeRSemble~\cite{Kirschstein2023NeRSemble}} and \textbf{K-hairstyle~\cite{Kim2021KHairstyle}}, to comprehensively evaluate the proposed method on both frontal and back-head reconstruction tasks. 

\begin{itemize}
    \item \textbf{NeRSemble~\cite{Kirschstein2023NeRSemble}} is a widely used benchmark for frontal face reconstruction. It provides high-quality multi-view video sequences specifically designed for 3D avatars. For each subject, a total of 11 video sequences are recorded, including four emotion (EMO) sequences, six expression (EXP) sequences, and one free performance (FREE) sequence. Each video frame captures 16 synchronized camera views distributed across approximately 120 degrees in the front. The evaluation follows two protocols: 1) \textbf{\emph{novel-view synthesis}}, where head poses and expressions from training sequences are used to render the subject from \textit{unseen camera viewpoints}. 2) \textbf{\emph{self-reenactment}}, where a held-out sequence with \textit{unseen poses and expressions} is used to drive the avatar, and rendering is performed across all 16 camera views.

    \item \textbf{K-hairstyle~\cite{Kim2021KHairstyle}} provides multi-view image sequences of multiple subjects, captured at 6-degree intervals across a full 360° azimuth range. 
    We utilize the back-view images (within $90^\circ$–$270^\circ$) as pseudo-ground-truth references to evaluate our synthesized results, addressing the lack of real supervisory signals in these regions. 
\end{itemize}

\begin{table*}[t]
\centering
\caption{Back-Head Region Evaluation. Quantitative Comparison of Baselines and AvatarBack Enhancements. $\mathbf{D}_{\textbf{GA}}$ and $\mathbf{D}_{\textbf{SF}}$ Denote the Subject Splits Following the GaussianAvatars and SurfHead Protocols, Respectively.\label{tab:combined_comparison}} 
\vspace{-5pt}
\begin{tabular}{l|ccc|cc}
\toprule
\textbf{Metric} 
& \multicolumn{3}{c|}{$\mathbf{D}_{\textbf{GA}}$} 
& \multicolumn{2}{c}{$\mathbf{D}_{\textbf{SF}}$}\\
\cmidrule(lr){2-4} \cmidrule(lr){5-6}
& PanoHead~\cite{Sizhe2023PanoHead}
& GaussianAvatars~\cite{Qian2024GaussianAvatars} & \cellcolor{gray!20}GaussianAvatars+AvatarBack & SurfHead~\cite{lee2025surfhead} & \cellcolor{gray!20}SurfHead+AvatarBack \\
\midrule
Clarity$\uparrow$ & 6.583 & 6.778 & \cellcolor{gray!20}8.444 & 7.014 & \cellcolor{gray!20}8.556 \\
Structural Integrity$\uparrow$ & 7.833 & 6.556 & \cellcolor{gray!20}8.306 & 6.653 & \cellcolor{gray!20}8.361 \\
Texture Quality$\uparrow$ & 6.319 & 6.000 & \cellcolor{gray!20}7.806 & 6.181 & \cellcolor{gray!20}8.125 \\
Color \& Lighting Consistency$\uparrow$ & 7.056 & 6.306 & \cellcolor{gray!20}8.181 & 7.125 & \cellcolor{gray!20}8.639 \\
Overall Perception$\uparrow$ & 6.917 & 6.375 & \cellcolor{gray!20}8.278 & 6.653 & \cellcolor{gray!20}8.444 \\
\midrule
Overall Score$\uparrow$ & 6.94 & 6.40 & \cellcolor{gray!20}\textbf{8.20} & 6.73 & \cellcolor{gray!20}\textbf{8.43} \\
\bottomrule
\end{tabular}
\vspace{-10pt}
\end{table*}
\begin{table}[t]
    \raggedright 
    \renewcommand{\arraystretch}{1.2}
    \setlength{\tabcolsep}{19pt}
    \caption{Back-Head Region Evaluation. Quantitative Comparison of Back-Head Reconstruction Quality on the K-Hairstyle Dataset.}
    \label{tab:back_metrics}
    \vspace{-5pt}
    \begin{tabular}{lcc}
        \toprule
        Model & FID$\downarrow$ & KID$\downarrow$ \\
        \midrule
        GaussianAvatars~\cite{Qian2024GaussianAvatars}           & 218.34 & 0.202 \\
        \rowcolor{gray!20}\textbf{GaussianAvatars+AvatarBack} & \textbf{146.73} & \textbf{0.120} \\
        \midrule
        SurfHead~\cite{lee2025surfhead}                  & 232.46 & 0.227 \\
        \rowcolor{gray!20}\textbf{SurfHead+AvatarBack}       & \textbf{165.06} & \textbf{0.146} \\
        \bottomrule
    \end{tabular}
    \vspace{-15pt}
\end{table}

\textbf{Criteria.}
\textbf{SurfHead+AvatarBack} follows the evaluation setup of SurfHead~\cite{lee2025surfhead}, where nine reconstructed avatars are selected for evaluation. Among these sequences, EMO-1 is used for self-reenactment evaluation, while the remaining eight sequences, excluding the FREE sequence, are used for training. For training, 15 out of the 16 camera views are used, excluding the 8th view. The 8th view, which corresponds to the central frontal camera, is reserved as the novel-view synthesis evaluation. \textbf{GaussianAvatars+AvatarBack} follows the evaluation protocol of GaussianAvatars~\cite{Qian2024GaussianAvatars}, using the same number of subjects and the same partitioning strategy. 
The specific subject IDs used in both protocols are provided in the Supplementary Material.

To comprehensively assess the quality of 3D avatar reconstruction, we employ different evaluation protocols for the frontal and back-head regions.

\begin{itemize}
    \item \textbf{Frontal Region Evaluation}: we adopt Peak Signal-to-Noise Ratio (PSNR), Structural Similarity Index Measure (SSIM), and Learned Perceptual Image Patch Similarity (LPIPS) to respectively assess pixel-level accuracy, structural consistency, and perceptual similarity. These metrics are consistent with the evaluation protocols of GaussianAvatars~\cite{Qian2024GaussianAvatars} and SurfHead~\cite{lee2025surfhead}, facilitating fair comparison. Among them, higher PSNR and SSIM values indicate better reconstruction quality, while lower LPIPS values correspond to higher perceptual similarity.
    
    \item \textbf{Distribution-Level Evaluation Metrics}: We evaluate the similarity between the distributions of reconstructed and reference images using Frechet Inception Distance (FID) and Kernel Inception Distance (KID), computed from Inception network~\cite{szegedy2015Inception} features. Lower values indicate better alignment. The evaluation compares synthesized images at azimuth angles from $90^\circ$ to $270^\circ$ across all time steps of nine subjects with corresponding reference images from the K-hairstyle dataset.
    
    \item \textbf{GPT-4o-based Perceptual Scoring}: To capture subtle perceptual differences aligned with human visual judgment, we adopt a GPT-4o-based~\cite{openai2024gpt4ocard} perceptual scoring system. The evaluation focuses on three representative back-head viewpoints: the exact rear view ($180^\circ$), the left-back view ($135^\circ$), and the right-back view ($225^\circ$). Each view is assigned a weight reflecting its contribution to overall reconstruction quality: $50\%$ for the rear view and $25\%$ for each side-back view. The final perceptual score $S$ is computed as a weighted sum:
    \begin{equation}
    S = 0.5 \times S_{180^\circ} + 0.25 \times S_{135^\circ} + 0.25 \times S_{225^\circ},
    \end{equation}
    where $S_{\theta}$ denotes the GPT-4o perceptual score at azimuth angle $\theta$.
    Each per-view score $S_{\theta}$ is computed based on five equally weighted criteria: clarity, structural integrity, texture quality, color and lighting consistency, and overall perception. 
    The detailed definitions of these criteria are provided in the Supplementary Material.
    
\end{itemize}

\begin{table}[!t]
    \centering
    \footnotesize
    \renewcommand{\arraystretch}{1.2}
    \setlength{\tabcolsep}{1pt}
    \caption{Frontal Region Evaluation on the NeRSemble Dataset. Top: Results Following the Evaluation Protocol of GaussianAvatars. Bottom: Results Following the Protocol of SurfHead.}
    \label{tab:combined_metrics}
    \vspace{-5pt}
    \begin{tabular}{c|ccc|ccc}
        \toprule
        \multirow{2}{*}{\textbf{Method}}  & \multicolumn{3}{c|}{Novel-View Synthesis} & \multicolumn{3}{c}{Self-Reenactment} \\
        \cmidrule(lr){2-4} \cmidrule(lr){5-7}
         & PSNR↑ & SSIM↑ & LPIPS↓ & PSNR↑ & SSIM↑ & LPIPS↓ \\
        \midrule
        AvatarMAV~\cite{Xu2023AvatarMAV} & 29.5 & 0.913 & 0.152 & 24.3 & 0.887 & 0.168 \\
        PointAvatar~\cite{Zheng20223PointAvatar} & 25.8 & 0.893 & 0.097 & 23.4 & 0.884 & 0.104 \\
        INSTA~\cite{Zielonka2023InstantVolumetricHead} & 26.7 & 0.899 & 0.122 & 26.3 & 0.906 & 0.110 \\
        GaussianAvatars~\cite{Qian2024GaussianAvatars} & 31.6 & 0.938 & 0.065 & 26.0 & 0.910 & 0.076 \\
        \rowcolor{gray!20}\textbf{\scriptsize  GaussianAvatars+AvatarBack} & \textbf{31.8} & \textbf{0.939} & \textbf{0.064} & \textbf{26.1} & \textbf{0.912} & \textbf{0.075} \\
        \midrule 
        PointAvatar~\cite{Zheng20223PointAvatar} & 20.56 & 0.844 & 0.206 & 20.59 & 0.854 & 0.190 \\
        Flare~\cite{Bharadwaj2023FLARE} & 21.91 & 0.814 & 0.228 & 21.11 & 0.802 & 0.227 \\
        SplattingAvatars~\cite{Shao2024SplattingAvatar} & 23.68 & 0.858 & 0.232 & 20.25 & 0.828 & 0.265 \\
        GaussianAvatars~\cite{Qian2024GaussianAvatars} & 30.29 & 0.934 & \textbf{0.067} & 23.43 & 0.891 & 0.093 \\
        SurfHead~\cite{lee2025surfhead} & 30.07 & 0.934 & 0.079 & 23.53 & 0.892 & 0.103 \\
        \rowcolor{gray!20}\textbf{SurfHead+AvatarBack} & \textbf{32.75} & \textbf{0.940} & 0.069 & \textbf{26.53} & \textbf{0.907} & \textbf{0.089} \\
        \bottomrule
    \end{tabular}
\end{table}
\begin{figure}[t]
    \centering
    \includegraphics[width=\linewidth]{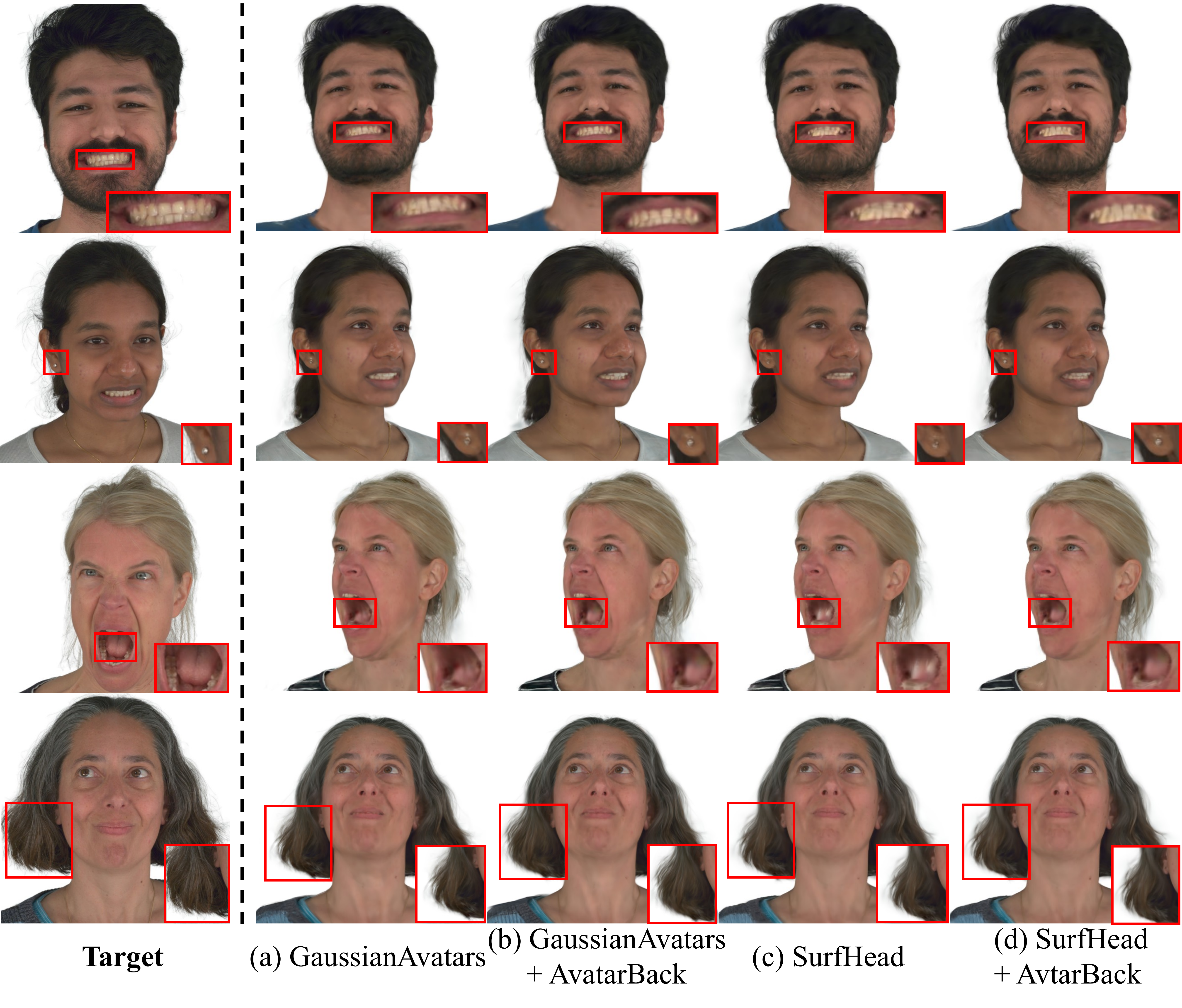} 
    \vspace{-20pt}
    \caption{Qualitative comparison of reconstructed frontal-views. The proposed AvatarBack method consistently enhances fine-grained details, leading to more faithful frontal reconstructions.}
    \label{fig:frontal_qualitative}
    \vspace{-20pt}
\end{figure}

\textbf{Baselines.}
We validate the proposed AvatarBack framework by integrating it with two state-of-the-art 3D head reconstruction methods,~\ie, GaussianAvatars~\cite{Qian2024GaussianAvatars} and SurfHead~\cite{lee2025surfhead}, yielding GaussianAvatars+AvatarBack and SurfHead+AvatarBack.
For frontal-view evaluations, we quantitatively compare the proposed method with the state-of-the-art facial reconstruction methods.
For back-head views, we propose to evaluate across multiple criteria.

\subsection{Quantitative Comparison}

\textbf{Back-head Region.}  
We assess the quality of the back-head regions with the cutting-edge GPT-4o model as shown in Tab.~\ref{tab:combined_comparison}.
It shows that AvatarBack obtains a profound enhancement in perceived visual quality.
The overall score for GaussianAvatars increases from 6.40 to 8.20, and for SurfHead, it rises from 6.73 to 8.43. 
The improvements are attributed to the enhancement of clarity, structural integrity, and texture quality.
PanoHead achieves a score of 6.94, higher than GaussianAvatars (6.40) but lower than the AvatarBack-enhanced methods (8.20). This shows that despite its multi-view generation capability, it still struggles to produce consistent and realistic back-head textures, underscoring the necessity of our framework.
Tab.~\ref{tab:back_metrics} further shows the evaluation results of the distribution statistics regarding the reconstructed back-head regions,~\ie, FID and KID.
Our framework yields remarkable improvements for both integrated methods. Specifically, GaussianAvatars+AvatarBack demonstrates a notable FID reduction of 71.61, and SurfHead+AvatarBack achieves a substantial decrease of 67.40. These consistent and considerable improvements in FID and KID metrics underscore that the synthesized textures are much closer to the distribution of real-world images, thereby enhancing visual fidelity and realism.
This qualitative assessment confirms that our framework effectively elevates the visual quality of back-head reconstructions.
\begin{figure*}[t]
\vspace{-15pt}
    \centering
    \includegraphics[width=\textwidth]{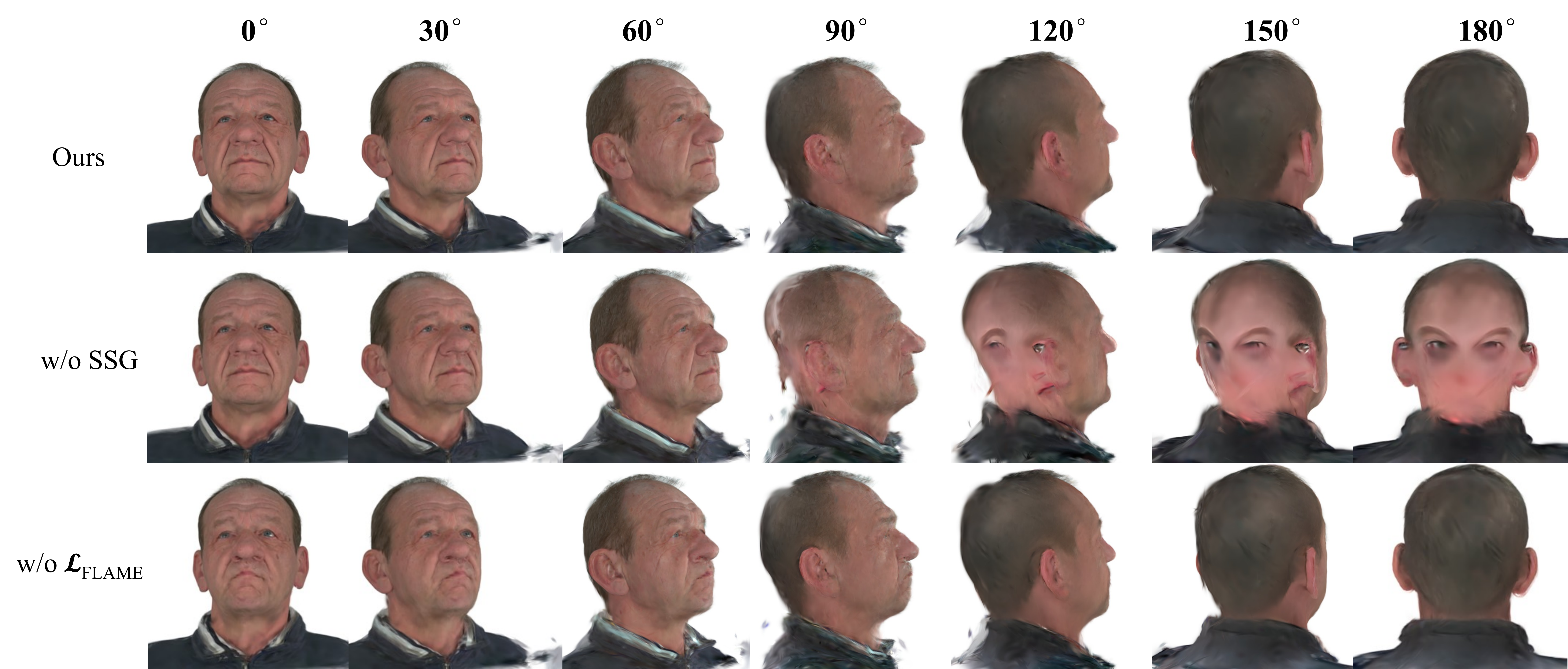}
    \vspace{-10pt}
    \caption{SSG improves back-head geometry, while $\mathcal{L}_{\text{flame}}$ preserves frontal facial structure and symmetry. The full model integrates both components to produce anatomically plausible and view-consistent 3D head reconstructions.}
    \label{fig:ablation_study}
    \vspace{-15pt}
\end{figure*}
\begin{figure}[!t]
\vspace{-5pt}
\centering
\includegraphics[width=\linewidth]{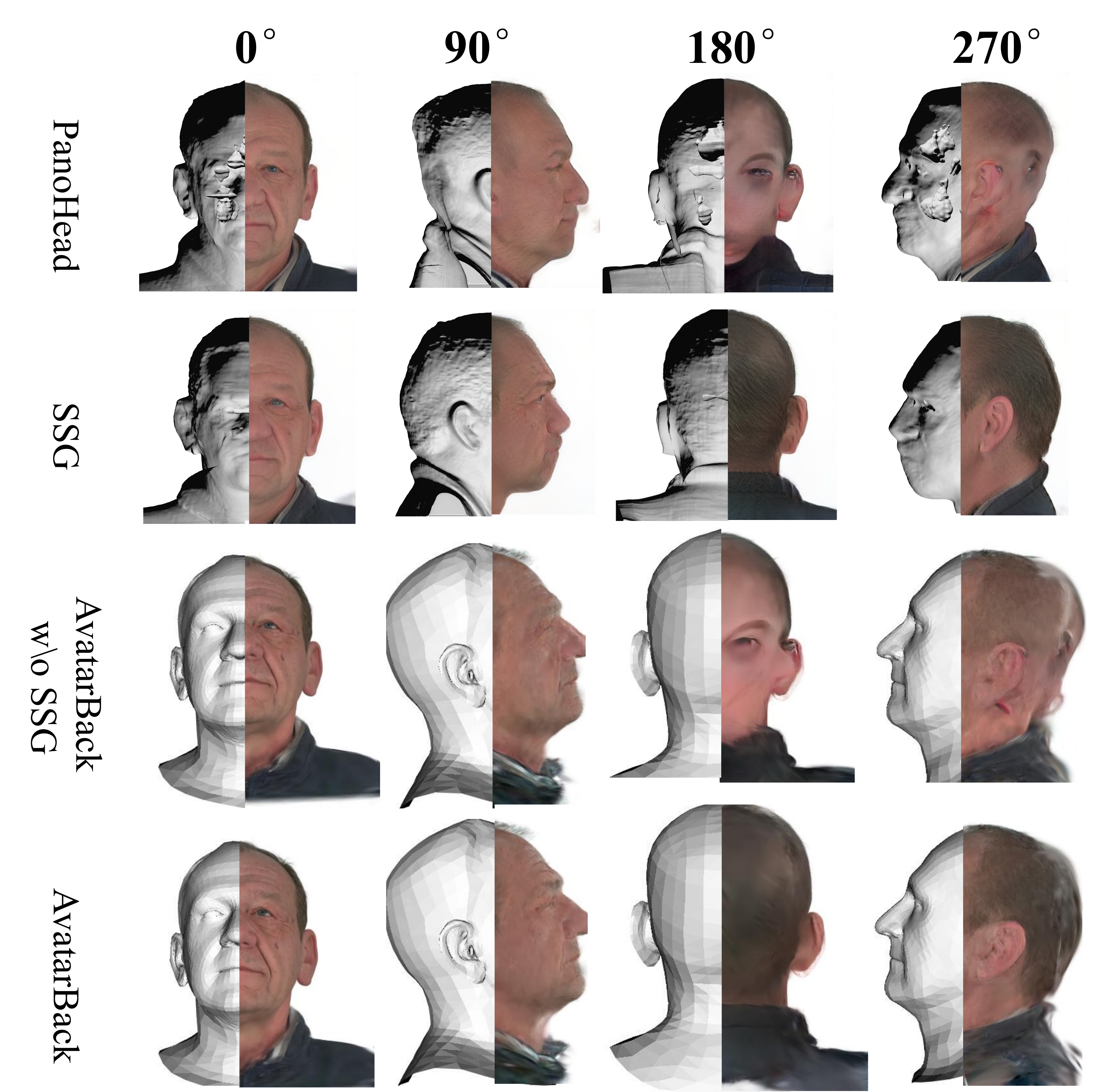} 
\vspace{-15pt}
\caption{Ablation comparison of mesh quality and rendered images under four canonical viewpoints. SSG enhances mesh completeness and fidelity, while AvatarBack enforces geometric regularity and recovers plausible back-head structures under reliable supervision.} 
\label{fig:mesh_comparison}
\vspace{-15pt}
\end{figure}

\textbf{Frontal Region.}
Tab.~\ref{tab:combined_metrics} presents a comprehensive quantitative comparison of our proposed AvatarBack framework against state-of-the-art 3D head reconstruction methods, rigorously assessing the quality of reconstructed frontal face regions. 
The results show that the proposed AvatarBack approach achieves competitive accuracy on both novel-view synthesis and self-reenactment tasks under multiple evaluation metrics.
Notably, SurfHead+AvatarBack delivers a substantial PSNR gain of 2.68 in novel-view synthesis and an even more significant increase of 2.99 in self-reenactment.
In the lower half of Tab.~\ref{tab:combined_metrics}, although GaussianAvatars records a marginally better LPIPS in the split following SurfHead, our approach exhibits the most balanced and comprehensively superior performance across all benchmarks.
These results confirm that our AvatarBack method not only effectively completes the unseen back-head regions but also maintains competitive reconstruction quality for the frontal face.

\subsection{Qualitative Comparison}

\textbf{Back-head Region.}  
Fig.~\ref{fig:backview_comparison_gs} provides a qualitative comparison of back-head reconstructions, contrasting the baseline GaussianAvatars and SurfHead models against their AvatarBack-enhanced counterparts. Each row visualizes outputs at challenging, unseen viewpoints ranging from $60^\circ$ to a full $180^\circ$.
The baseline methods, trained solely on frontal-facing data, exhibit a progressive decline in reconstruction quality as the viewpoint shifts toward the back. Around $90^\circ$, noticeable hollow regions begin to appear in the reconstructed geometry. At $180^\circ$, corresponding to the direct back view, the reconstructions collapse entirely, manifesting as disorganized floating artifacts and incoherent structures, rather than resembling a plausible head or hair configuration. 
In stark contrast, the models enhanced with AvatarBack seamlessly address these deficiencies. They successfully fill the geometric void with dense, plausible hair volume that maintains the subject's overall head shape and hairstyle. Even at a full $180^\circ$ posterior view, the models render a solid and continuous surface. The synthesized texture is not only internally coherent but also integrates naturally with the visible hair from the original views, creating a complete and visually convincing 360-degree appearance. These results suggest that AvatarBack significantly enhances the visual realism and structural integrity of reconstructions under unseen back-view conditions.

\textbf{Frontal Region.}  
Fig.~\ref{fig:frontal_qualitative} presents frontal-view comparisons based on GaussianAvatars~\cite{Qian2024GaussianAvatars} and SurfHead~\cite{lee2025surfhead}. 
In the first row, AvatarBack contributes to more defined tooth structures, particularly in (b) and (d), compared to the baseline outputs (a) and (c). The second row highlights improved reconstruction of earrings, with clearer and sharper results in the enhanced versions. The third row shows better definition around the mouth corners, especially when comparing (b) to (a). In the fourth row, hair details are more faithfully reproduced with AvatarBack. 
These results indicate that our AvatarBack not only significantly enhances the quality of local facial features but also ensures consistent improvements, thereby elevating overall structural and perceptual fidelity.

\subsection{Ablation Studies}
We evaluate the two core components of AvatarBack: the subject-specific generator (SSG) and FLAME-based regularization $\mathcal{L}_{\text{flame}}$.  
SSG generates reliable back-head geometry, mitigating semantic leakage from single-view training, while $\mathcal{L}_{\text{flame}}$ preserves frontal facial structure and symmetry.  
Removing either module leads to distorted or incomplete reconstructions, whereas the full model produces anatomically plausible and view-consistent 3D heads.

\textbf{Analysis of Mesh Quality.}  
We qualitatively compare reconstructed meshes and rendered images across four canonical viewpoints ($0^\circ$, $90^\circ$, $180^\circ$, $270^\circ$) as shown in Fig.~\ref{fig:mesh_comparison}.  
PanoHead exhibits severe surface irregularities, profile distortions, and back-view artifacts, including misreconstructed facial features.  
SSG improves mesh completeness and fidelity across views but still lacks structured regularity.  
AvatarBack w/o SSG introduces some geometric regularity, yet inaccurate back-head supervision results in implausible geometry.   
Our full AvatarBack pipeline, guided by SSG, produces smooth surfaces, realistic facial structures, and consistent head geometry across all viewpoints.

Additional ablation results, including the effects of the FLAME-based regularization term, are provided in the Supplementary Material.

\section{Conclusion}
In this paper, we proposed AvatarBack, a plug-and-play framework that leverages generative supervision and adaptive spatial alignment to complete the missing back-head regions in Gaussian-based 3D head avatars. Our method achieves geometrically consistent, photorealistic, and fully animatable reconstructions, while surpassing existing approaches in both fidelity and completeness. This work paves the way toward more complete, realistic, and controllable 3D human head modeling.

\bibliographystyle{IEEEtran}
\bibliography{paper}

\end{document}


\maketitle

\section{Additional Experimental Details}
\textbf{Evaluation Protocol Details.}
Table~\ref{tab:ners_eval_subjects} lists the specific subject IDs used in both protocols on the NeRSemble dataset. As described in the main paper, $\mathbf{D}_{\textbf{GA}}$ and $\mathbf{D}_{\textbf{SF}}$ denote the subject selections following the protocols of GaussianAvatars~\cite{Qian2024GaussianAvatars} and SurfHead~\cite{lee2025surfhead}, respectively.

\begin{table}[ht]

  \renewcommand{\arraystretch}{1.3} 
  \setlength{\tabcolsep}{5.7pt} 
  \centering
  \small
   \caption{Subject IDs Used for Evaluation on the NeRSemble Dataset.}
   \label{tab:ners_eval_subjects}
  \begin{tabular}{lccccccccc}
    \hline
    \textbf{$\mathbf{D}_{\textbf{GA}}$} & 074 & 104 & 218 & 253 & 264 & 302 & 304 & 306 & 460 \\
    \textbf{$\mathbf{D}_{\textbf{SF}}$}        & 074 & 140 & 175 & 210 & 253 & 264 & 302 & 304 & 306 \\
    \hline
  \end{tabular}

\end{table}

\textbf{Details of GPT-4o-based Perceptual Scoring.}
Table~\ref{tab:backhead_gpt4o_criteria} summarizes the five criteria used by GPT-4o to assess the quality of back-of-head reconstruction. Each criterion contributes equally (20\%) to the per-view score $S_\theta$, which is then aggregated into the overall perceptual score $S$ as described in the main paper.

\begin{table}[ht]
\setlength{\tabcolsep}{6pt} 
\renewcommand{\arraystretch}{1.2}
\caption{Evaluation Criteria Used by GPT-4o to Assess the Quality of Back-Head Reconstruction.} 
\label{tab:backhead_gpt4o_criteria}
\centering
\begin{tabularx}{\linewidth}{@{}>{\bfseries}l X@{}}
\toprule
\textbf{Criterion} & \textbf{Evaluation Description} \\
\midrule
Clarity & \textit{Are the reconstructed details (~\eg, hair, skin, contours) sharp and clear? Is the visual clarity comparable to the frontal view?} \\
\midrule
Structural Integrity & \textit{Does the back-of-head structure align with the subject’s frontal structure? Are proportions and symmetry preserved?} \\
\midrule
Texture Quality & \textit{Are hair and skin textures realistic and detailed? Do they match the visual style of the input image?} \\
\midrule
Color \& Lighting Consistency & \textit{Are skin and hair color consistent with the front image? Is the lighting coherent with the frontal view?} \\
\midrule
Overall Perception & \textit{Does the image appear realistic and natural overall? Is it visually consistent with the front-view reference?} \\
\bottomrule
\end{tabularx}
\vspace{-10pt}
\end{table}

\section{Additional Ablation Studies}
\begin{figure}[htbp]
\centering
\includegraphics[width=\linewidth]{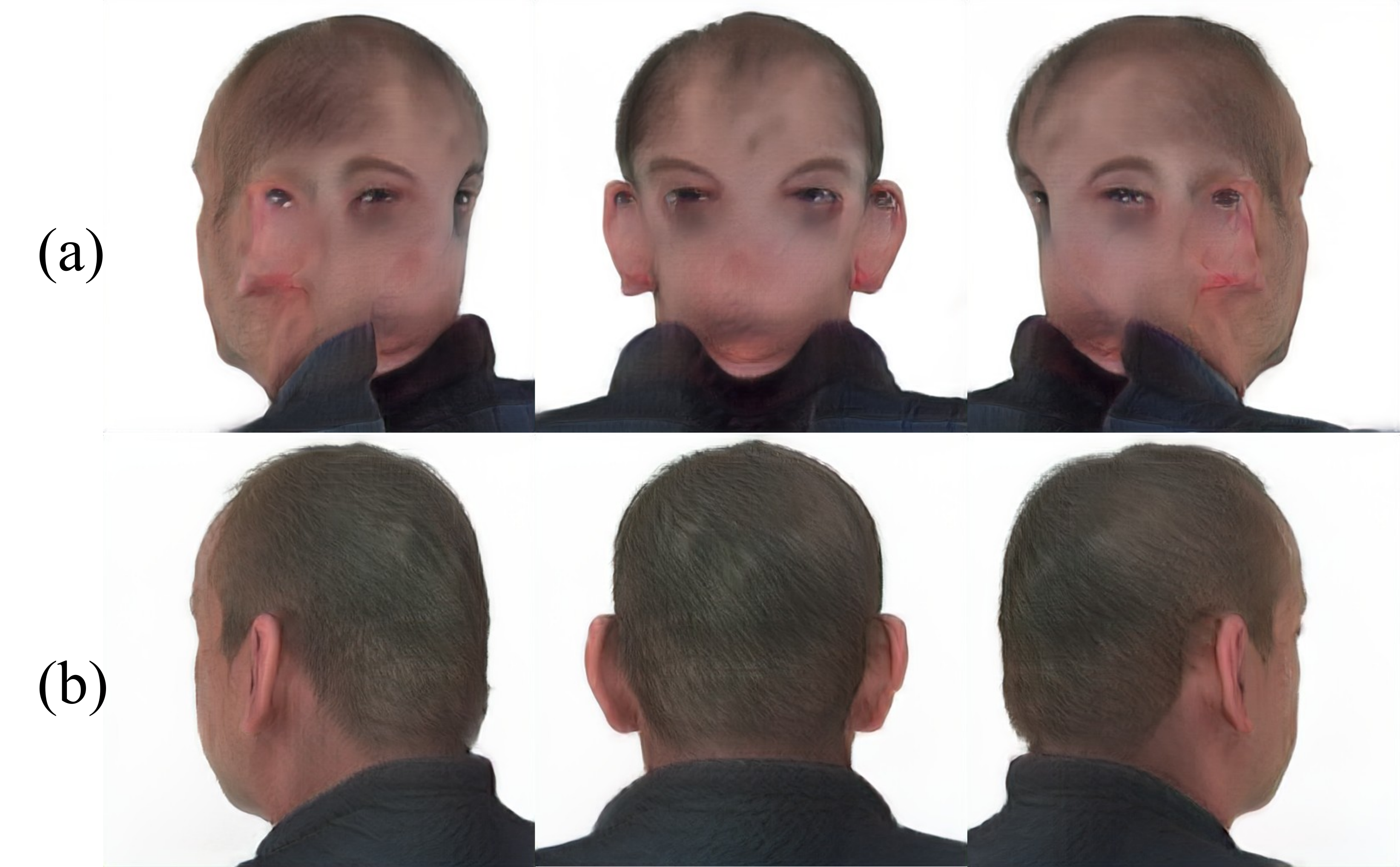}
\caption{Qualitative comparison of back-head reconstruction using PanoHead with single-view input versus our multi-view training strategy. (a) Results using only one frontal image. (b) Results using multiple frontal views (Ours). Our approach produces more accurate geometry and detailed textures in the back-head region.}
\label{fig:panohead+PTIvsOur}
\end{figure}

\textbf{Subject-specific Generator Module.}
Notably, as illustrated in Fig.~\ref{fig:panohead+PTIvsOur}(a), when the generator module is trained using only a single frontal image, the reconstructed back-head often contains implausible artifacts, including front-facing features such as eyes, eyebrows, and nose. This indicates an over-reliance on visible cues, resulting in semantic leakage and poor generalization to occluded areas.
When such flawed pseudo images are used to guide GaussianAvatars+AvatarBack, the resulting 3DGS head avatar exhibit incomplete geometry and unrealistic back-head appearance. In contrast, our SSG generates more reliable and view-consistent geometry shown in Fig.~\ref{fig:panohead+PTIvsOur}(b), which effectively supports AvatarBack in producing accurate and coherent head reconstructions.

\begin{figure}[htbp]
\centering
\includegraphics[width=\linewidth]{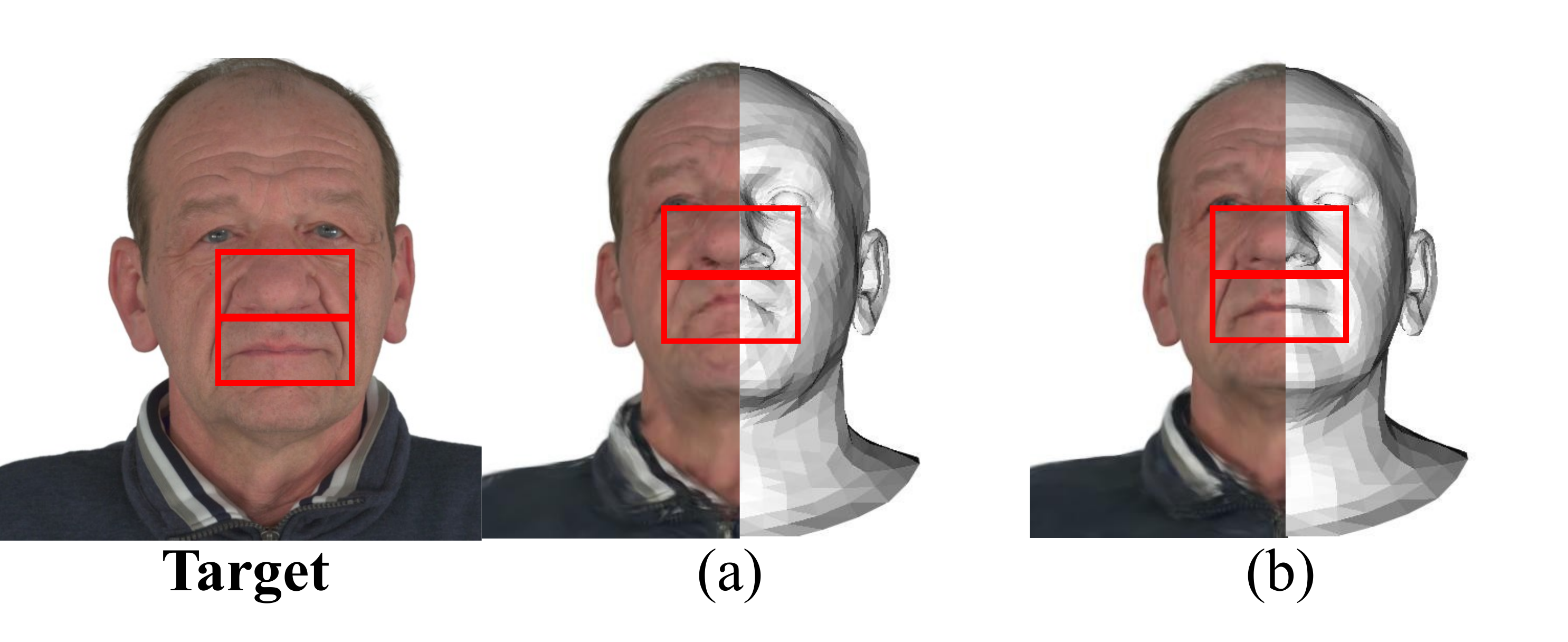}
\caption{Effect of FLAME-based regularization. From left to right: Target, (a) mesh without $\mathcal{L}_{\text{flame}}$, and (b) mesh with $\mathcal{L}_{\text{flame}}$. The regularized mesh (b) exhibits improved frontal structure and better alignment with true facial geometry.}
\label{fig:flame_regularization}
\end{figure}

\textbf{FLAME-based Regularization.}
We evaluate the effect of the FLAME-based regularization term $\mathcal{L}_{\text{flame}}$, which constrains mesh deformation to ensure plausible and stable facial geometry. As shown in Fig.~\ref{fig:flame_regularization}, (a) the mesh without this regularization exhibits obvious distortion in the frontal region, such as unnatural facial contours and collapsed structures. In contrast, (b) the mesh with $\mathcal{L}_{\text{flame}}$ more closely resembles the ground-truth results, preserving facial symmetry and anatomical plausibility.
This effect is also reflected in the full reconstruction pipeline, where omitting $\mathcal{L}_{\text{flame}}$ leads to visibly distorted features in the frontal view. Compared to the full model, these results highlight the importance of the FLAME-based prior in enforcing geometric realism and preventing frontal-view collapse.

\bibliographystyle{IEEEtran}
\bibliography{supp}